\documentclass{article}
\usepackage{spconf,amsmath,epsfig,xcolor,booktabs,blindtext,subcaption}
\usepackage{siunitx}
\DeclareGraphicsExtensions{.pdf,.jpg,.png,.jpeg}
\graphicspath{{images/}, {figs/}}

\newcommand{\figref}[1]{Figure~\ref{fig:#1}}
\newcommand{\tabref}[1]{Table~\ref{tab:#1}}

\title{Remote Estimation of Free-Flow Speeds}
\name{Weilian Song, Tawfiq Salem, Hunter Blanton, Nathan Jacobs}
\address{
  Department of Computer Science, University of Kentucky, USA\\
  \{weilian,salem,hunter,jacobs\}@cs.uky.edu
}

\begin{document}
%
\maketitle
\begin{abstract}
  We propose an automated method to estimate a road segment's free-flow speed from overhead imagery and road metadata. The free-flow speed of a road segment is the average observed vehicle speed in ideal conditions, without congestion or adverse weather. Standard practice for estimating free-flow speeds depends on several road attributes, including grade, curve, and width of the right of way. Unfortunately, many of these fine-grained labels are not always readily available and are costly to manually annotate. To compensate, our model uses a small, easy to obtain subset of road features along with aerial imagery to directly estimate free-flow speed with a deep convolutional neural network (CNN). We evaluate our approach on a large dataset, and demonstrate that using imagery alone performs nearly as well as the road features and that the combination of imagery with road features leads to the highest accuracy.
  
\end{abstract}


\section{Introduction}

The behavior of an average automobile driver is based on numerous factors. In this work, we focus on one aspect of behavior: the speed of travel. We specifically focus on the free-flow speed, which is the average vehicle speed along a roadway when there is no congestion or adverse weather conditions. The free-flow speed is used in a wide variety of planning and regulatory contexts. Standard practice for estimating free-flow speeds relies on the availability of many road features, such as lane widths and curvature. While some of these features are easy to gather, others require expensive surveying equipment and expert annotators, restricting their availability. As a result, in many states, roads are assigned the default speed limit of 35 mph for urban areas and 55 mph for rural areas, which can often be inconsistent with actual operational speeds and lead to undesired driver behavior.

To automate the costly process of estimating free-flow speeds, we propose a deep convolutional neural network (CNN) that takes as input the overhead imagery of the roadway and coarse-grained road features, including the posted speed limit, the functional classification, and the type of area. As output, our CNN generates a probability distribution over integer free-flow speeds. For training data, we use aggregated data collected from real-world drivers. At inference time, since the input features are all easily obtainable, it should be possible to rapidly, and inexpensively, estimate free-flow speeds over large spatial regions.

Using a large-scale evaluation dataset, we find that using only image features result in the poorest performance, but that combining imagery and road features result in significantly better performance than road features alone.  

\begin{figure}
  \centering
  \includegraphics[width=\linewidth]{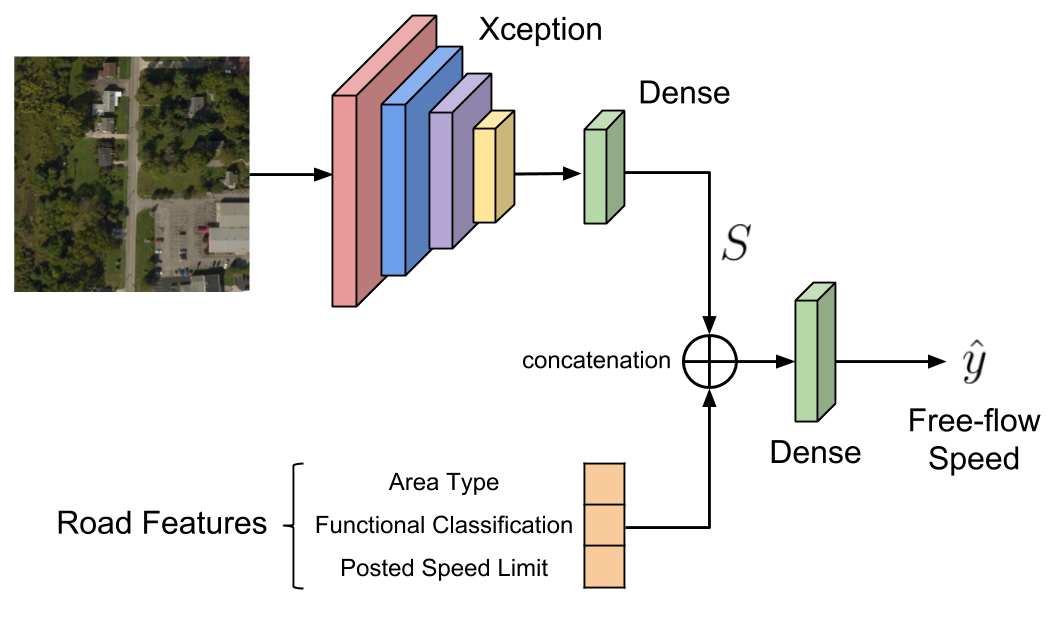}
  \caption{Network architecture for predicting free-flow speeds.}
  \label{fig:network}
\end{figure}

\begin{figure*}[ht]
  \centering
  \includegraphics[width=\textwidth]{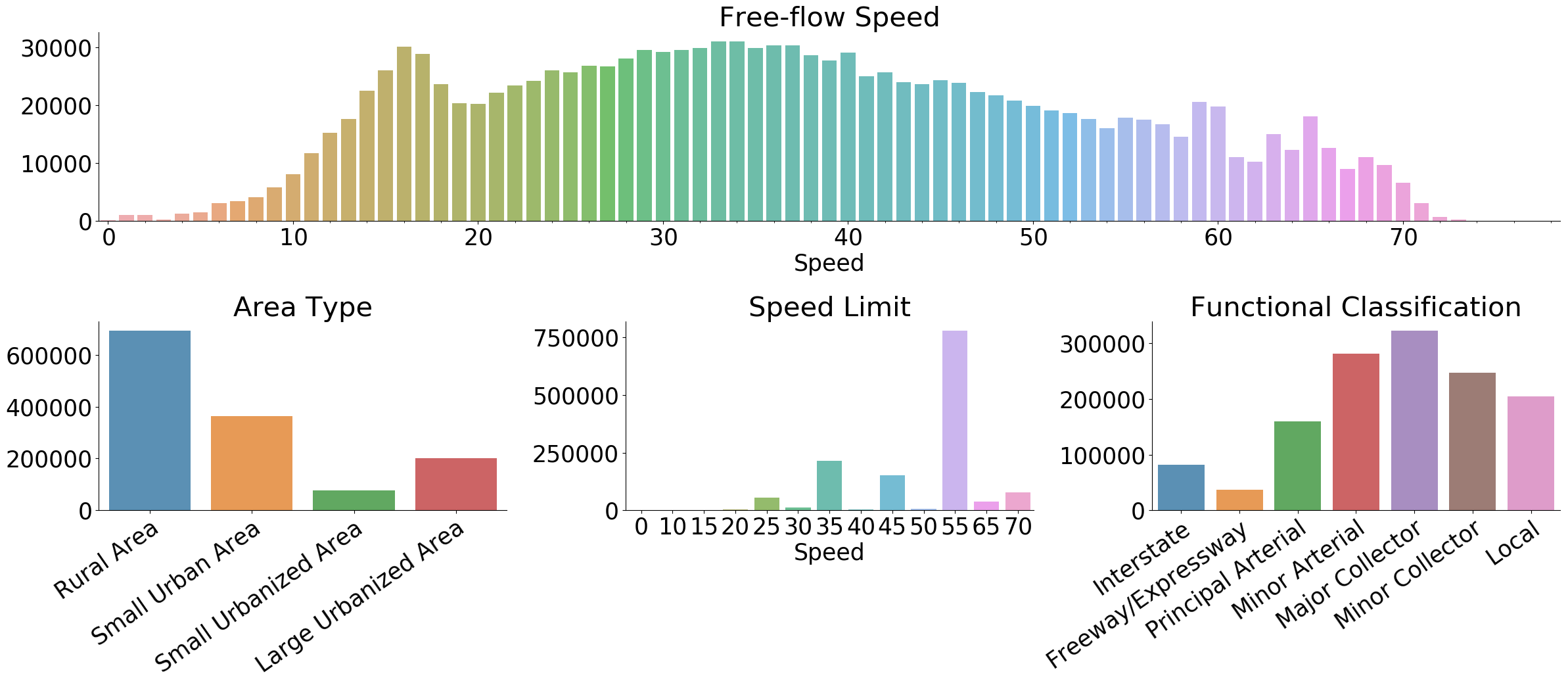}
  \caption{Histograms of various road features in our dataset.}
  \label{fig:distribution}
\end{figure*}

\section{Related Work}
Different studies have been proposed to estimate and map properties of the visual world using overhead images. Several authors have proposed different deep learning based approaches for vehicle detection~\cite{sakla2017deep,sommer2017fast} and road extraction~\cite{zhang2018road, bastani2018roadtracer, alshehhi2017hierarchical, mattyus2017deeproadmapper} from aerial images. Salem et al.~\cite{salem2018soundscape} introduced an approach for mapping soundscapes of geographic regions using overhead imagery. Greenwell et al.~\cite{greenwell2018objects} proposed a model that is capable of predicting object histograms from overhead imagery. Several works have addressed the problem of speed estimation. Shuai et al.~\cite{hua2018vehicle} introduced an approach to estimate the vehicle speed in traffic videos. Most similar to our work, Song et al.~\cite{song2018farsa} proposed a model for road safety estimation based on the usRAP Star Rating Protocol. While this star rating is based on approximately 60 road safety features~\cite{harwood2010validation}, their network works directly on ground level panorama images. We propose a new method that instead uses overhead imagery with the addition of auxiliary features.

\section{Approach}
\label{sec:approach}
We utilize a CNN architecture to estimate the free-flow speed of a given road segment. The neural network uses both aerial imagery and relevant road features as input, and outputs a probability mass function over $K$ possible free-flow speeds.  We begin by describing the dataset that we use, followed by a more detailed description of the proposed network architecture.

\subsection{Dataset}
\label{ssec:dataset}
Our free-flow speed dataset is obtained through HERE Technologies and further annotated to incorporate coarse-grained road feature data needed for training. To calculate the free-flow speed for a particular road segment, driving speeds of vehicles were monitored during the year 2014. As free-flow speed refers to the speed that a driver can achieve without traffic congestion, we only consider data during non-holiday weekday periods from 9am to 3pm. The speeds are then averaged to obtain the ground-truth for each road segment. We rounded each ground-truth speed to the nearest integer to obtain a discrete label for training. This results in $K=79$ unique free-flow speeds.

For each road segment, we obtain an aerial image through the National Agriculture Imagery Program (NAIP) data. Each image is centered at the beginning of the segment, and covers an aerial view of $400\times 400$ m\textsuperscript{2} of land. We orient the images such that the direction of travel is up, and we reshape it to $224\times 224$. In total, our dataset consists of \num{1335132} road segments and their corresponding aerial images.

\figref{distribution} shows the distribution of the free-flow speeds along with road features used for training. The drastic difference between the distribution of free-flow speeds and speed limits is evident here. In fact, over half of the roads have free-flow speeds significantly slower than the speed limits (more than 10-miles-per-hour). We will examine a few roads with such differences in Section~\ref{ssec:qual_eval}, using predictions obtained from our model.

We split our dataset into training and testing sets. To ensure the sets are disjoint, we partition the dataset by location. The dataset is divided into counties, eight of which are selected for evaluation. The resulting test set is roughly $7\%$ of the total data, and contains a variety of roads with different area types, terrain, and free-flow speeds. A validation set ($1\%$ of the training set) is reserved for model selection.

\subsection{Network Architecture}
\label{ssec:network_arch}
The backbone of our architecture is the Xception~\cite{chollet2016xception} model, which is used to extract high-level features from the aerial image. The resulting feature vector of length $2048$ is passed to a dense layer with output size $512$, denoted as $S$, as shown in~\figref{network}.
In addition to aerial imagery, the network is trained using three integer features related to driving speed: area type, functional classification, and posted speed limit. The three features are concatenated with $S$ before feeding the combined feature vector into
the final dense layer with output size $K$. We denote the output of the final layer as $\hat{y}$.

\subsection{Loss Function}
\label{ssec:loss}
We formulate the task of predicting the free-flow speed of a road segment as a multiclass classification problem with ground-truth free-flow speed label $l$. With the output of the network defined in \ref{ssec:network_arch}, we compute the training loss as the standard cross-entropy loss between the predicted distribution $\hat{y}$ and the target distribution $y$, as shown in the following equation:
\begin{equation}
  L=-\frac{1}{N}\sum_{i=1}^N\sum_{k=1}^K y_{i,k}\log \hat{y}_{i,k}
\end{equation}
where $N$ is the number of training examples.

\subsection{Implementation Details}
\label{ssec:imple_details}
Our model is implemented using Tensorflow. The network is optimized using the Adam optimizer with default parameters. We initialize the Xception network with weights pre-trained on ImageNet for image classification. During training, we freeze the weights of the Xception network and optimize only the last two dense layers with learning rate $0.001$. We decay the learning rate exponentially by a factor of 10 every 5 epochs. {\em ReLU} activation layers are used throughout the network except for the last layer, which uses the Softmax activation instead. We apply $L_2$ regularization to the two dense layers with scale $0.00005$. We train with batch size 16, for a total of 15 epochs.
\begin{table}[]
  \centering
  \caption{Within-5 accuracy for each method.}
  \begin{tabular}{lc}
    \toprule
    \multicolumn{1}{c}{Method} & \multicolumn{1}{c}{Within-5 Acc.} \\ \hline
    Imagery Only       & 37.60 \\
    Road Features Only & 40.07 \\
    Combined           & \textbf{49.86} \\
    \bottomrule
  \end{tabular}
  \label{tab:eval}
\end{table}

\begin{figure}
\centering
    \begin{subfigure}{.45\linewidth}
        \centering
        \includegraphics[width=0.98\linewidth]{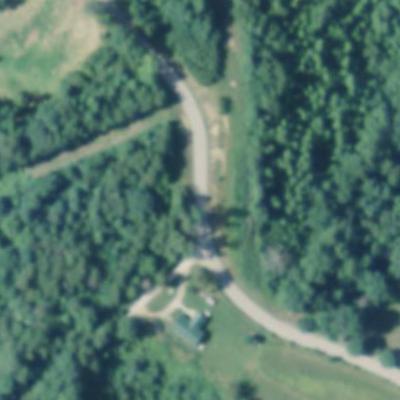}
        \caption{Free-flow: 18, Limit: 55}
    \end{subfigure}%
    \begin{subfigure}{.45\linewidth}
        \centering
        \includegraphics[width=0.98\linewidth]{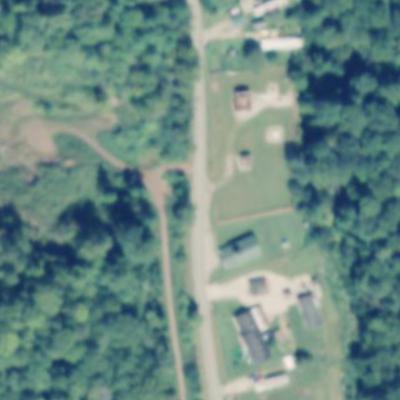}
        \caption{Free-flow: 17, Limit: 55}
    \end{subfigure}
    
    \begin{subfigure}{.45\linewidth}
        \centering
        \includegraphics[width=0.98\linewidth]{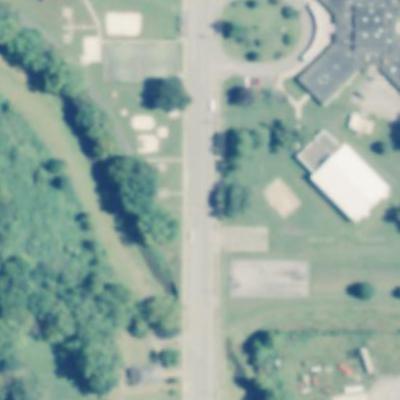}
        \caption{Free-flow: 27, Limit: 15}
    \end{subfigure}%
    \begin{subfigure}{.45\linewidth}
        \centering
        \includegraphics[width=0.98\linewidth]{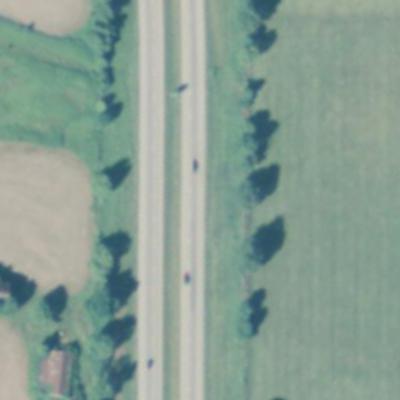}
        \caption{Free-flow: 67, Limit: 55}
    \end{subfigure}
    \caption{Examples of discrepancies in the predicted free-flow speed and the speed limit of roads in miles per hour. Predictions obtained from the combined feature model.}
    \label{fig:discrep}
\end{figure}

\begin{figure}[t!]
\centering
    \begin{subfigure}{.45\linewidth}
        \centering
        \includegraphics[width=0.98\linewidth]{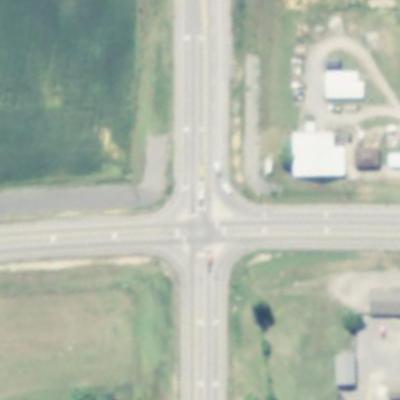}
        \caption*{method A: 50, method B: 16}
    \end{subfigure}%
    \begin{subfigure}{.45\linewidth}
        \centering
        \includegraphics[width=0.98\linewidth]{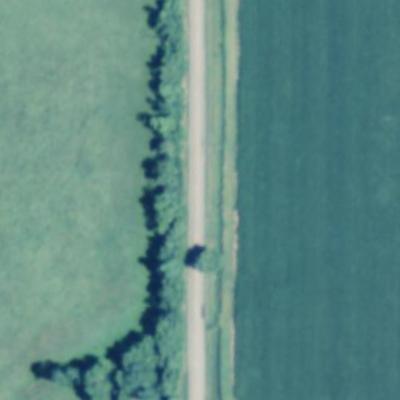}
        \caption*{method A: 31, method B: 42}
    \end{subfigure}
    \hfill
    \begin{subfigure}{.45\linewidth}
        \centering
        \includegraphics[width=0.98\linewidth]{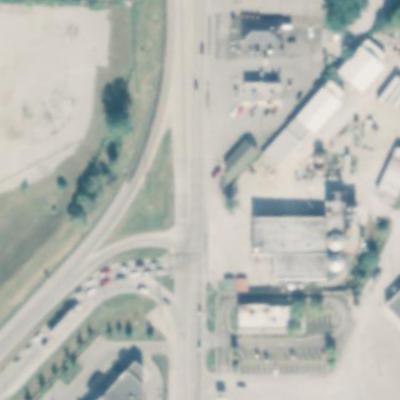}
        \caption*{method A: 46, method B: 15}
    \end{subfigure}%
    \begin{subfigure}{.45\linewidth}
        \centering
        \includegraphics[width=0.98\linewidth]{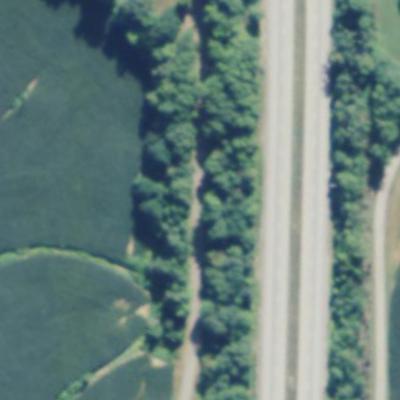}
        \caption*{method A: 24, method B: 50}
    \end{subfigure}
    \hfill
    \begin{subfigure}{.45\linewidth}
        \centering
        \includegraphics[width=0.98\linewidth]{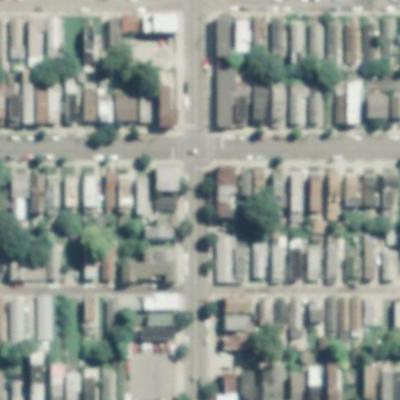}
        \caption*{method A: 30, method B: 15}
    \end{subfigure}%
    \begin{subfigure}{.45\linewidth}
        \centering
        \includegraphics[width=0.98\linewidth]{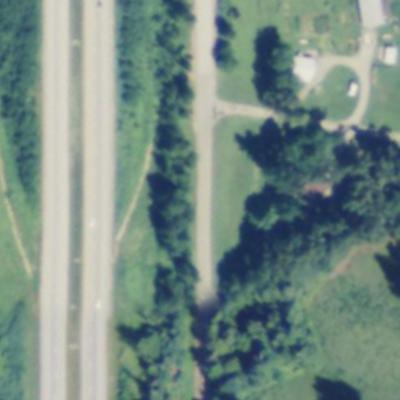}
        \caption*{method A: 24, method B: 40}
    \end{subfigure}
    \caption{Free-flow speed predictions in miles per hour from two methods, where method A is the road feature only model and method B is the combined feature model.}
    \label{fig:versus}
\end{figure}

\section{Evaluation}
\label{sec:eval}
Using the dataset described in \ref{ssec:dataset}, we trained our proposed model along with two other variations: imagery-only model and road-features-only model. We conducted quantitative and qualitative evaluation on our reserved test set.

\subsection{Quantitative Analysis}
\label{ssec:quant_eval}
We aim to discover the effect of each input modality in predicting free-flow speeds. We calculate the accuracy of the three trained models (overhead imagery only, road features only, and both) on the test set. \tabref{eval} displays the within-5 accuracy for each method, where we consider a prediction to be positive if it is within five-miles-per-hour of the ground-truth speed. As we can see, road feature data is better than aerial imagery for predicting free-flow speeds, but we obtain the best performance when we combine both modalities.

\subsection{Qualitative Evaluation}
\label{ssec:qual_eval}
Free-flow speed should be similar to the speed limit for a given road, but that is often not the case. We are interested in finding roads where the predicted free-flow speed is drastically different than the speed limit. \figref{discrep} shows such examples. The top two roads have free-flow speeds much lower than the speed limits, most likely due to the curvature and number of intersections present. The bottom two roads have free-flow speeds much higher than the speed limits, which is understandable since the roads are straight, wide, and without congestion. Detection of these discrepancies is very useful, as traffic engineers can quickly filter through millions of roads to identify ones that need to have their speed limits re-evaluated.

We also compared predictions from our top two models: one trained on both aerial imagery and road features, and the other trained only on road features. \figref{versus} shows the images that had drastically different predictions from the two models. The images in the right column received higher free-flow speed predictions from the combined model, where the images in the left column received higher free-flow speed predictions from the feature only model. These differences further support the importance of image features for prediction, as imagery can provide fine-grained information about the road segment. For the left three roads, it is natural for people to drive slower in residential areas or at intersections; for the right three, straight roads and ones parallel to interstates allow drivers to achieve greater speeds. Fine-grained labels such as presence of intersections, population density, and curvature of the road are hard to obtain, but they can easily be inferred from aerial imagery.

\section{Conclusion}
\label{sec:conclusion}

We introduced a method for estimating the free-flow speed of a road segment, which is important for understanding driver behavior. We demonstrated that a combination of aerial imagery and related road features as input is best for prediction, since it obtained higher accuracy than models trained on either features alone. We also performed qualitative evaluation, and obtained insights on the effect of input modalities along with the relationship between free-flow speeds and speed limits. We hope to extend this work and include ground-level imagery as an additional input, since information such as roadside hazards may not be visible in aerial imagery.

\section*{Acknowledgements}

We thank Mei Chen and Xu Zhang, both from the Univ.\ of Kentucky Dept.\ of Civil Engineering, for their assistance in obtaining and understanding this dataset. We also acknowledge the support of NSF CAREER (IIS-1553116).

{\small
\bibliographystyle{IEEEbib}
\bibliography{biblio}
}

\end{document}